# Learning Entropy and Spatial Adaptation Dynamics of Multilayer Perceptrons for Structural Point Extraction

Jan Glaser[a], Ivo Bukovsky[*a,b], Marcel Jirina[a]

[a]*Czech Technical University in Prague, Prague, Czech Republic*
[b]*University of South Bohemia in Ceske Budejovice, Ceske Budejovice, Czech Republic*
[*]*Corresponding author*

**Abstract**

This paper extends the concept of Learning Entropy (LE) from temporal adaptive systems to spatial learning in multilayer perceptron networks (MLPs) applied to image data. Instead of evaluating image structure directly from gradients or covariance operators, as local neighborhood methods do, the proposed approach analyzes the learning process itself through Learning Entropy. An MLP is trained to predict the intensity of a center pixel from its surrounding spatial context, while LE is evaluated from the incremental adaptation of neural weights during learning across image-derived samples. The resulting Spatial Learning Entropy Maps (SLEM) identify unusual image points and regions that induce strong adaptation of the neural network and therefore have an important role in the learning process. The results indicate that spatial Learning Entropy provides a complementary perspective to conventional feature extraction and explainability methods by highlighting spatial locations that are particularly informative for network learning. Spatial Learning Entropy provides a complementary perspective to conventional feature extraction and explainability methods by identifying image points and regions according to their learning impact rather than their local structural properties. The proposed framework may open new directions for learning-driven image or scene analysis in computer vision, manufacturing, and robotics.

## 1. Introduction

Feature extraction and structural point detection belong among the fundamental problems of computer vision and image analysis. Classical approaches such as the Harris detector [1], Shi–Tomasi detector [2], and Canny edge detector [3] evaluate local image structure using spatial gradients, covariance matrices, or edge responses. These methods identify geometrically significant image regions and are widely used in tracking, navigation, image registration, and scene analysis. Their underlying assumption is that image structure can be estimated directly from local geometric properties of the image itself.

With the development of deep learning systems, feature extraction has increasingly shifted toward learned latent representations obtained from neural networks trained directly on image data. Convolutional neural networks introduced by LeCun et al. [4] demonstrated that hierarchical neural representations can significantly outperform handcrafted image operators in many visual tasks. Modern deep-learning systems therefore often rely on learned internal representations rather than explicitly designed image descriptors [5].

Increasing complexity of neural architectures has simultaneously led to growing interest in explainability and interpretability of neural systems. Methods such as saliency maps [6], Integrated Gradients [7], Grad-CAM [8], LIME [9], and SHAP [10] attempt to identify which input regions contribute most strongly to prediction outputs of trained neural networks. These approaches primarily analyze inference sensitivity, i.e., the relationship

$$\left[\mathbf{x}(k), y(k)\right] \rightarrow \tilde{y}(k)\,, \tag{1}$$

where $\mathbf{x}(k)$ denotes the input vector, $y(k)$ denotes the desired output, and $\tilde{y}(k)$ denotes the neural prediction.

However, inference sensitivity does not necessarily characterize adaptation behavior of the learning system itself. The presented work is motivated by the hypothesis that structurally important information may also emerge from adaptation dynamics during learning. Instead of evaluating which image regions contribute to neural predictions, the proposed approach evaluates which image regions induce strong adaptation activity in the neural model itself.

The underlying idea is closely related to adaptive systems and incremental learning theory. In adaptive filters and online learning systems, information about observed dynamics is inherently encoded in parameter adaptation trajectories [11], [12]. Samples inducing strong parameter updates correspond to regions where the current model is insufficiently adapted to observed data. Such adaptation, i.e., learning activity therefore carries information about novelty, governing dynamics, and structural complexity from the viewpoint of the learning system itself.

This perspective is related to continual learning [13], neural plasticity [14], novelty detection [15], and online adaptation [16], where adaptation trajectories themselves become objects of analysis. Unlike conventional probabilistic information measures, adaptation-based quantities depend not only on statistical occurrence of data, but also on the internal state of the learning system and its currently learned representation of governing laws.

Learning Entropy (LE) was originally introduced as a non-probabilistic information measure for adaptive filters and incremental learning systems [17], [18], [19]. Unlike Shannon entropy, which evaluates information according to probability distributions, Learning Entropy evaluates novelty according to adaptation activity of the learning system itself. In the presented work, the multilayer perceptron is trained incrementally to predict the intensity value of a center pixel from its surrounding local neighborhood in image. Each neighborhood represents an individual training sample which induces a neural adaptation response. The magnitude of the resulting weight adaptation may then be associated with the spatial position of the corresponding center pixel within the image.

Let $\Delta\mathbf{w}(k)$ denote the parameter update vector induced by the $k$-th training sample. The corresponding learning information [18] may generally be expressed as

$$L(k) = f\left(\Delta\mathbf{w}(k)\right) \tag{2}$$

where $f(\cdot)$ denotes a function quantifying adaptation activity. Large parameter updates correspond to observations producing strong adaptation response of the learning system, while small updates indicate observations already well represented by the current neural model.

Previous Learning Entropy [17], [18], [19] research focused mainly on temporal adaptive systems and in-parameter-linear neural architectures such as Higher-Order Neural Units. In such systems, parameter adaptation dynamics can often be analyzed directly because of favorable convergence and stability properties. However, image and point-cloud data introduce substantially more complex spatial structures. In these cases, polynomial in-parameter-linear architectures become computationally inefficient due to rapidly growing parameter dimensionality. Multilayer perceptrons therefore represent a natural extension toward more expressive nonlinear representations capable of modeling higher-dimensional spatial relationships.

The presented work investigates whether adaptation dynamics of multilayer perceptrons contain spatial structural information about image data itself. Instead of constructing feature maps directly from image gradients or covariance matrices, the proposed approach analyzes spatial distributions of neural adaptation activity during sample-wise learning. The proposed framework therefore evaluates the relationship

$$\left[\mathbf{x}(k), y(k)\right] \rightarrow \Delta\mathbf{w}(k) \tag{3}$$

where $\Delta\mathbf{w}(k)$ denotes the neural weight updates induced by the current observation training pair. The resulting Spatial Learning Entropy Maps (SLEM), that we introduce in this paper, may therefore be interpreted as adaptation sensitivity maps rather than conventional inference-attribution maps. In this context, adaptation sensitivity refers to the magnitude of the neural model's weight adaptation induced by individual samples during incremental learning. The objective of the paper is not to propose another conventional feature detector competitive with existing computer vision operators. Instead, the presented work investigates whether learning dynamics themselves may serve as a meaningful representation of spatial structure and adaptation-induced information in image data.

The results indicate that statistical complexity and adaptation-induced complexity may represent fundamentally different properties of image data, suggesting a possible connection between adaptive learning systems, continual learning, and learning-based information measures.

After the Introduction, Section 2 recalls the concept of Learning Entropy and extends it newly for spatial adaptation dynamics of MLP, Section 3 introduces the Spatial Learning Entropy Maps, Section 4 contains the experimental analysis with discussions, and Section 5 concludes the paper.

## 2. Learning Entropy and Spatial Adaptation Dynamics

Classical Shannon self-information is defined as

$$I(\mathbf{x}) = -\log_2 p(\mathbf{x}), \tag{4}$$

where $p(\mathbf{x})$ denotes the probability of occurrence of sample $\mathbf{x}$. The Shannon entropy therefore evaluates information according to statistical occurrence of data.

However, probabilistic rarity does not necessarily correspond to novelty from the perspective of an adaptive learning system. A data sample may be statistically rare while still conforming to an already learned governing law. Conversely, statistically common samples may induce strong adaptation activity when governing dynamics change.

Based on learning information defined in Eq. (2), the Learning Entropy can be estimated directly from neural parameter updates as

$$LE(k) = ||\Delta\mathbf{w}(k)||_1, \tag{5}$$

where $||\cdot||_1$ denotes the $L_1$ norm. Newly in images, the Learning Entropy therefore can quantify adaptation activity induced by local image neighborhoods during sample-wise learning as follows. Local image neighborhoods contain spatial information about edges, corners and intensity transitions surrounding the observed center pixel. In this study, the neighborhoods are presented to the MLP sequentially in a row-wise raster scan order. If the center pixel value can be accurately predicted from its neighborhood, the image neighborhood is typically structurally regular and well represented by the current neural model. Conversely, neighborhoods inducing strong neural adaptation indicate spatial regions whose structure is more difficult for the model to predict. The proposed approach evaluates image structure through adaptation dynamics during neighborhood-based prediction learning.

Let the grayscale image be represented as

$$X \in [0,1]^{H \times W}, \tag{6}$$

where $H$ and $W$ denote image height and width. For every pixel position $(i,j)$, a local neighborhood of radius $r$ is extracted. The corresponding training pair is defined as

$$(\mathbf{x}_{i,j}, y_{i,j}), \tag{7}$$

where

$$\mathbf{x}_{i,j} = \{X(u,v)\ ;\ (u,v) \in N_r(i,j)\ ;\ (u,v) \neq (i,j)\}, \tag{8}$$

where $N_r(i,j)$ denotes the local neighborhood of radius $r$ around coordinates $(i,j)$, and

$$y_{i,j} = X(i,j) \tag{9}$$

is the training target, i.e., the intensity of pixel at coordinates $(i,j)$.

The learning system is represented by a perceptron network with one hidden layer as follows

$$\tilde{y}_{i,j} = \sigma\left(\mathbf{w_2} \cdot \mathrm{ReLU}\left(\mathbf{W_1} \cdot x_{i,j}\right)\right) \tag{10}$$

where $\mathbf{W}_1$ is the hidden layer weight matrix, $\mathbf{w}_2$ are output neuron weights, and $\boldsymbol{\sigma}(\cdot)$ denotes the sigmoid activation (vector) function.

The MLP is trained incrementally in a sample-wise manner. For every image local neighborhood, used as an input sample, the corresponding parameter updates are recorded and assigned to the spatial position of the neighborhood center pixel. The accumulated adaptation responses across all image neighborhoods form a Spatial Learning Entropy Map $LE(i,j)$ which represents adaptation sensitivity of the neural model across the image domain (see Fig. 1 (b)). The evolution of these maps across training epochs is illustrated in Fig. 3.

Structurally complex image regions induce larger Learning Entropy values and stronger neural adaptation activity, while homogeneous regions generally produce smaller parameter changes.

The resulting spatial LE maps therefore represent spatial learning dynamics of the neural network rather than conventional edge maps.

The proposed approach therefore evaluates the relationship of feature vector and target value as in Eq. (3), rather than the conventional inference relationship

$$\mathbf{x}(k) \rightarrow y(k). \tag{11}$$

## 3. Spatial Learning Entropy Maps and Adaptation Activity

The Spatial Learning Entropy Map is proposed to identify structurally important image regions. A subset of significant points is obtained by thresholding

$$S_{LE}=\{(i,j);LE(i,j)\geq\tau\}, \tag{12}$$

where $\tau$ denotes a threshold parameter.

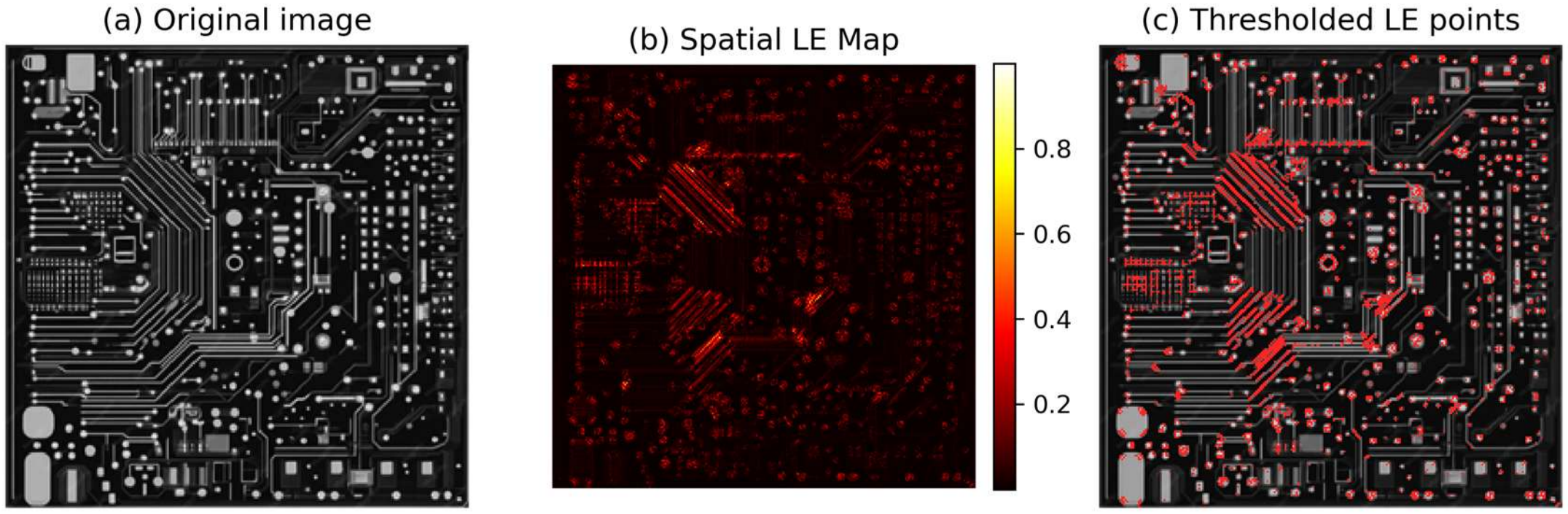

*Fig. 1: Learning Entropy-based point extraction pipeline. (a) the original image, (b) Spatial Learning Entropy Map, (c) the thresholded LE points.*

To quantify the total adaptation effort of the neural network during optimization (learning), cumulative adaptation activity is introduced by

$$A=\sum_{e=1}^{E}\sum_{l=1}^{L}\|\mathbf{w}_l^{(e)}-\mathbf{w}_l^{(e-1)}\|_F, \tag{13}$$

where $E$ denotes the number of training epochs, $L$ denotes the number of trainable layers, and $||\cdot||_F$ denotes the Frobenius norm. For a matrix $B\in\mathbb{R}^{m\times n}$, the Frobenius norm is defined as

$$||\ B\ ||_F=\sqrt{\sum_{i=1}^{m}\sum_{j=1}^{n}b_{i,j}^2}\ . \tag{16}$$

The quantity $A$, defined in Eq. (13) represents cumulative adaptation activity during optimization and therefore quantifies the total adaptation effort of the neural model.

## 4. Experimental Analysis and Discussion

The proposed LE-based point extraction was experimentally compared with the Harris corner detector, Shi–Tomasi corner detector and the Canny edge detector.

Experiments were performed on multiple synthetic geometries and industrial images. The synthetic datasets, including lines, circle, square, triangle, star, grid and gradient were intentionally selected because they allow controlled analysis of spatial adaptation dynamics. Industrial images, including circuit image, industrial fasteners, gears and a sample from the CAMEL Dataset for Visual and Thermal Infrared Multiple Object Detection and Tracking [20] (sequence 3, frame 000001) were selected to examine whether the observed adaptation characteristics generalize beyond idealized geometric primitives. The complete set of evaluated images along with the LE-selected points is shown in Fig. 5 and Fig. 6.

The Table 1 and Fig. 2 summarize the mean cumulative adaptation activity $\overline{A}$ obtained from repeated 100 independent training runs together with the corresponding Shannon entropy values for individual images. Here, $\overline{A}$ denotes the mean of the cumulative adaptation activity $A$ across the 100 runs.

Table 1: Mean cumulative adaptation activity obtained from 100 independent training runs together with the corresponding Shannon entropy values for individual images.

| Image | $\overline{A}$ | Shannon entropy |
|---|---|---|
| Synthetic geometric structures | | |
| lines | 0.694534 | 0.347379 |
| circle | 2.430559 | 0.997587 |
| square | 2.856517 | 0.944114 |
| triangle | 1.688779 | 0.825794 |
| star | 1.909273 | 1.088602 |
| grid | 1.133081 | 0.831609 |
| gradient | 1.252901 | 7.018223 |
| Industrial images | | |
| circuit | 0.926658 | 6.574729 |
| industrial fasteners | 0.626966 | 4.067724 |
| gears | 0.551749 | 4.055549 |
| CAMEL-Seq3-Frame1 [20] | 0.532140 | 7.276387 |

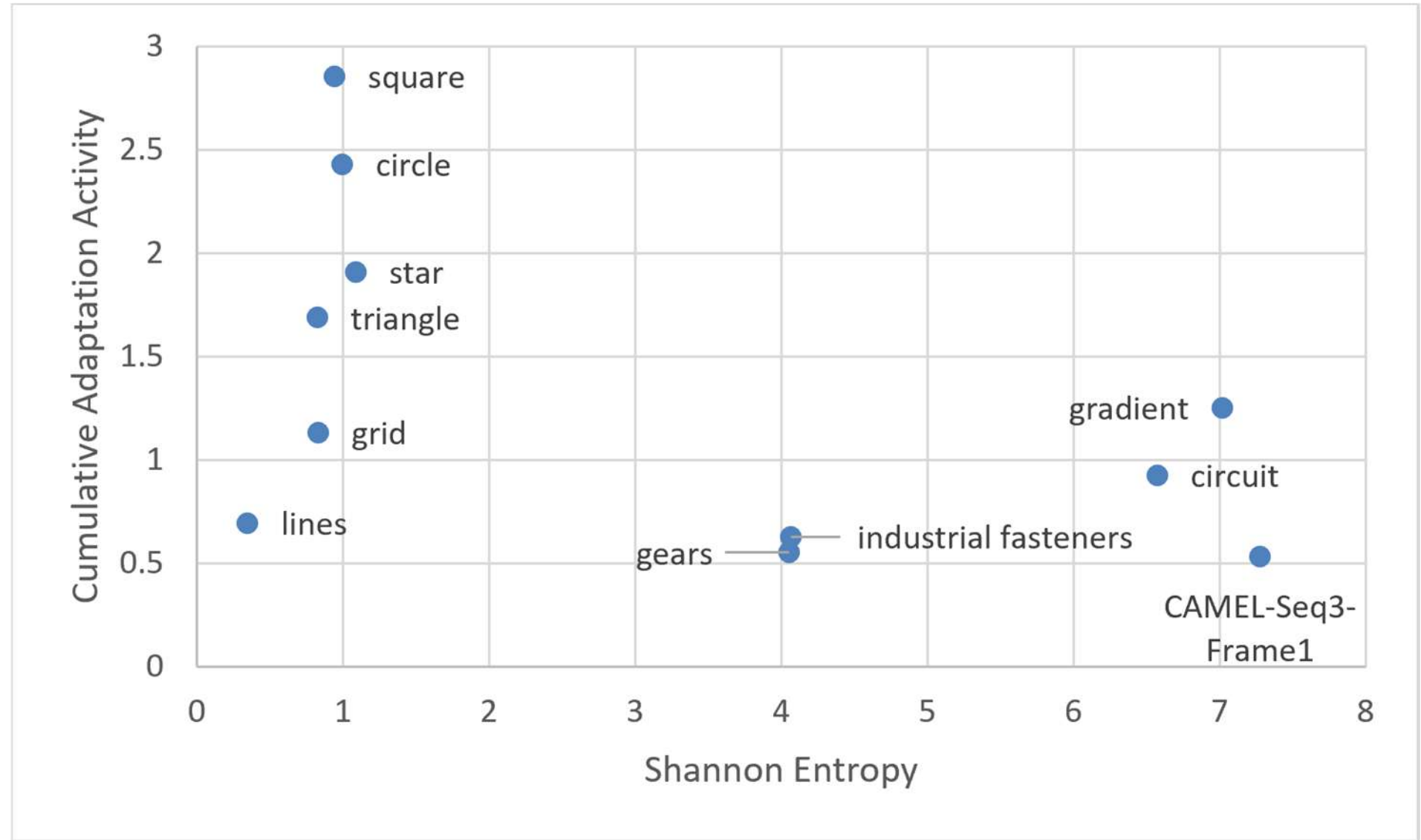

Fig. 2: The relationship between Shannon entropy and cumulative adaptation activity displaying its ability to distinguish among images, while Shannon entropy is rather similar (points on the left). The reported cumulative adaptation activity corresponds to the mean value computed on 100 independent runs.

The circuit and the gradient images are particularly important because it exhibits large Shannon entropy while not producing proportionally large adaptation activity. The observation suggests that statistical complexity and adaptation-induced complexity are fundamentally different quantities.

The resulting spatial LE maps consistently reveal edges, corners, intersections, and structurally complex image regions from the viewpoint of neural adaptation. The extracted LE points visually correspond to meaningful image structures similarly to classical feature detectors (see Fig. 7). However, the proposed method fundamentally differs from explainability approaches such as SHAP.

Classical explainability methods primarily analyze inference sensitivity, i.e., the contribution of input features to the output of an already trained model. In contrast, the proposed framework analyzes adaptation dynamics during learning itself. The extracted image regions are therefore important not necessarily because they dominate prediction outputs, but because they strongly influence evolution of the internal learning state.

Repeated independent experiments with random neural initializations were performed in order to evaluate statistical stability of the adaptation activity measure. The resulting analysis indicates that the adaptation activity and Shannon entropy are fundamentally different quantities.

An important observation is that the Learning Entropy concept survives the transition from in-parameter-linear neural architectures [21] toward multilayer perceptrons. Previous Learning Entropy research relied heavily on Higher-Order Neural Units because of their in-parameter-linearity, analyzability, and stability properties. However, image data introduce substantially more complex spatial structures for which polynomial architectures become computationally inefficient.

The presented results therefore suggest that Learning Entropy may represent a more general property of adaptation dynamics rather than a property tied to a specific neural architecture. Increased LE values are consistently observed around structurally significant image regions such as corners and edges. This shows partial spatial correspondence with classical feature extraction methods, including Shi–Tomasi, Harris, and Canny detectors (as seen in Fig. 7).

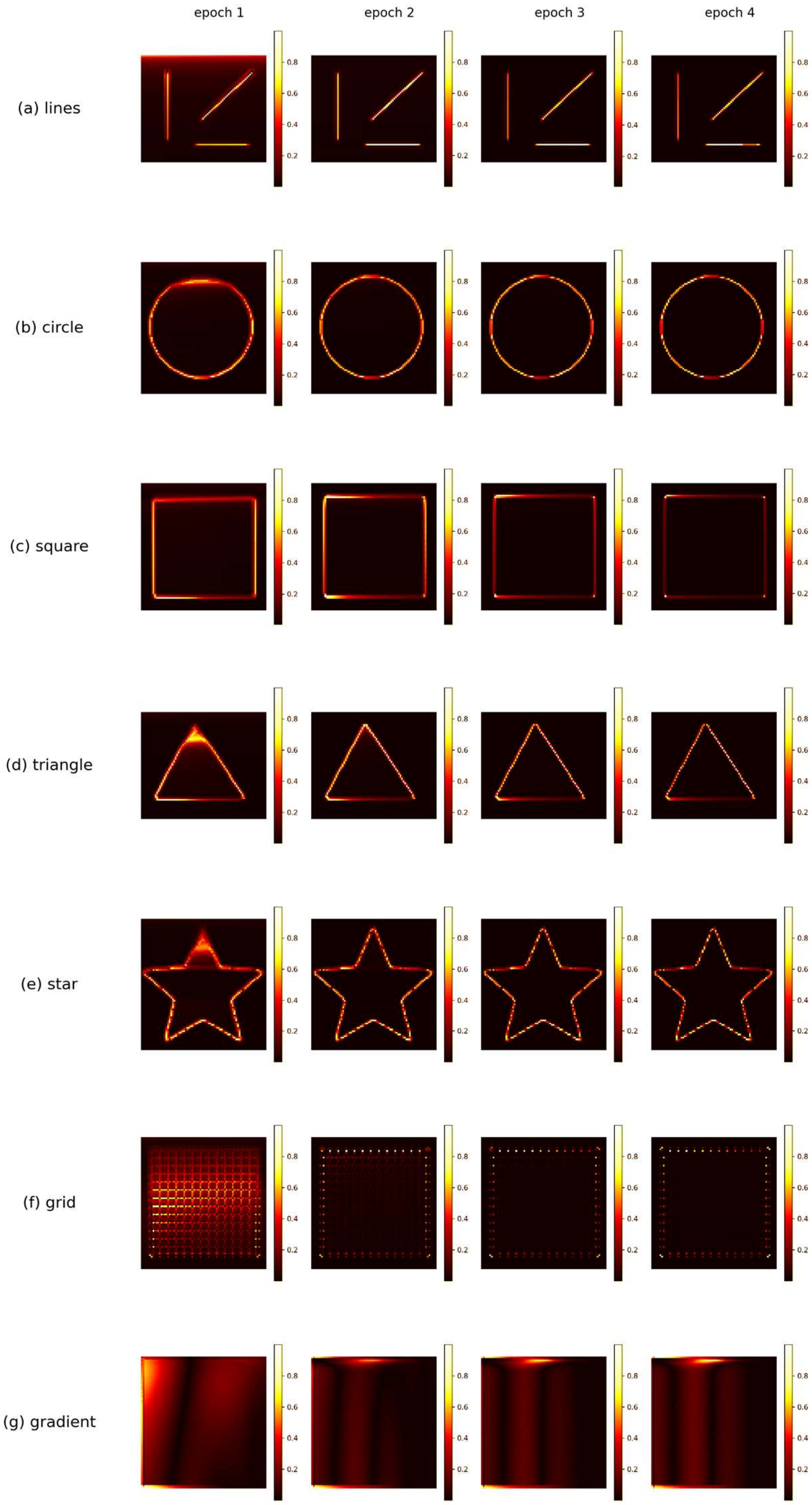

*Fig. 3 Evolution of spatial LE maps representing neural adaptation activity during sample-wise learning across epochs for synthetic geometric structures. The progression shows the stabilization of regions that dominate parameter updates during learning for various images.*

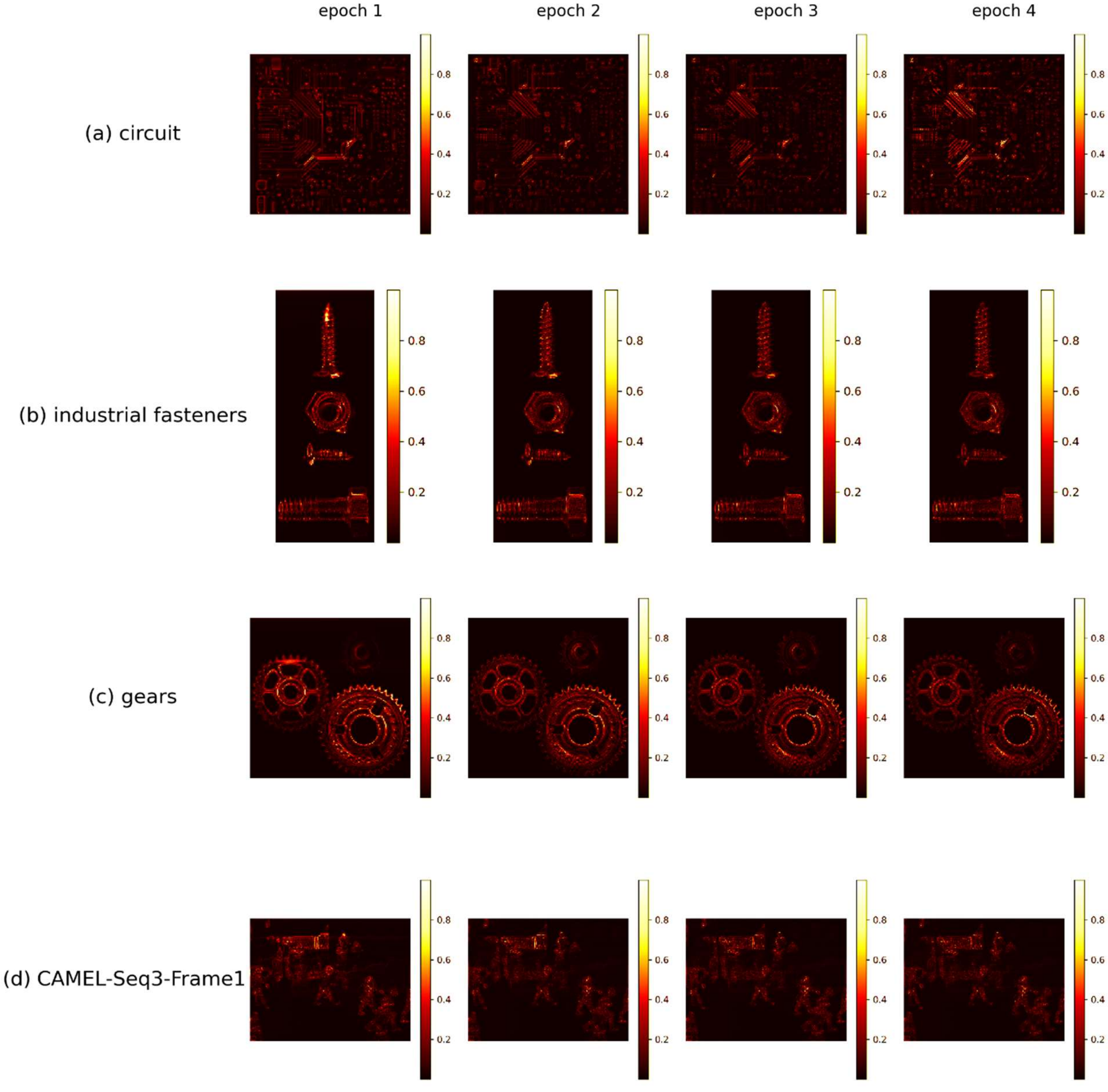

*Fig. 4: Evolution of spatial LE maps representing neural adaptation activity during sample-wise learning across epochs for industrial images. The progression shows the stabilization of regions that dominate parameter updates during learning for various images.*

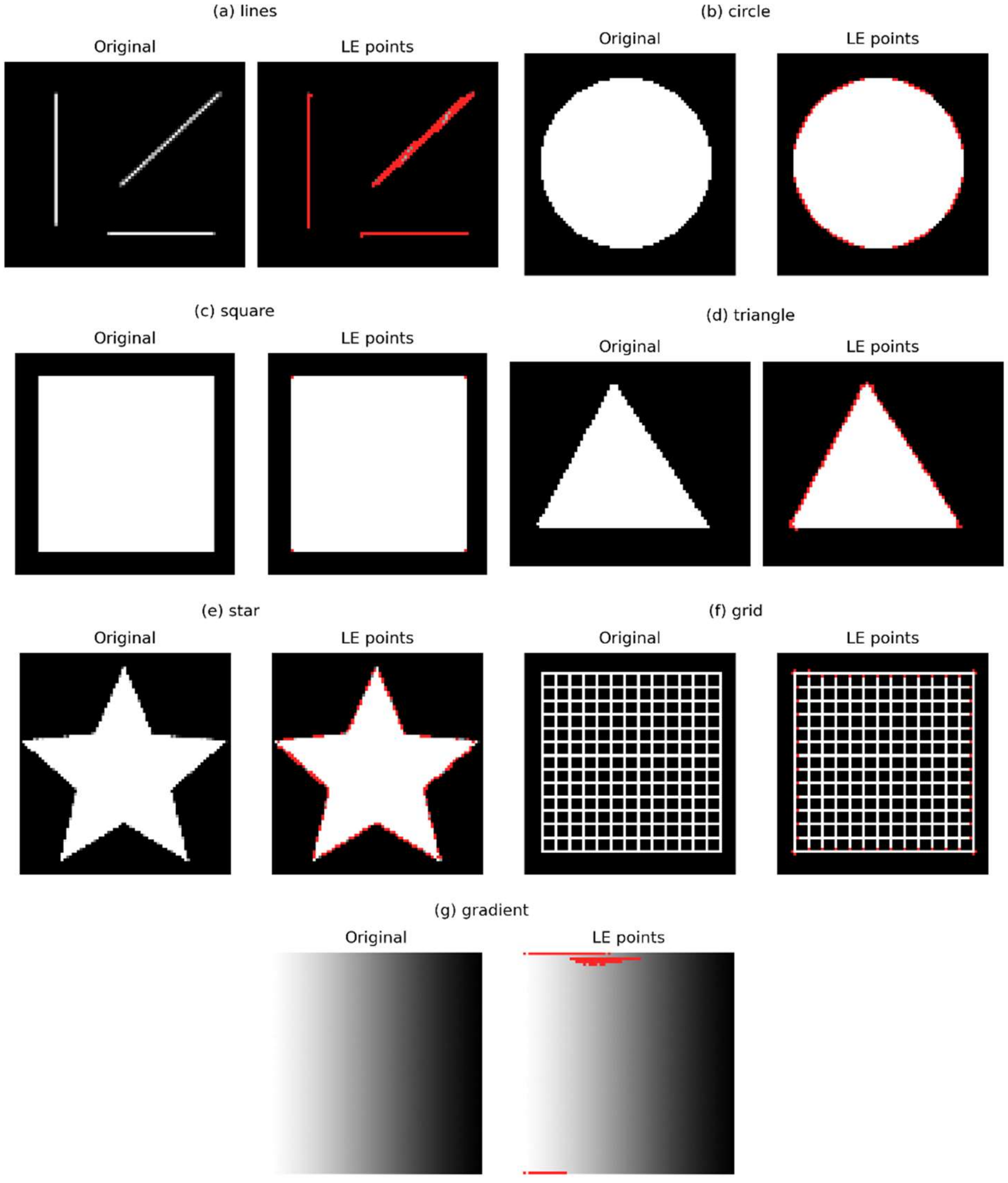

*Fig. 5: Spatial distribution of LE-selected points across synthetic geometric structures. The highlighted points denote pixel locations associated with high neural adaptation responses during incremental sample-wise learning, showing structurally informative regions.*

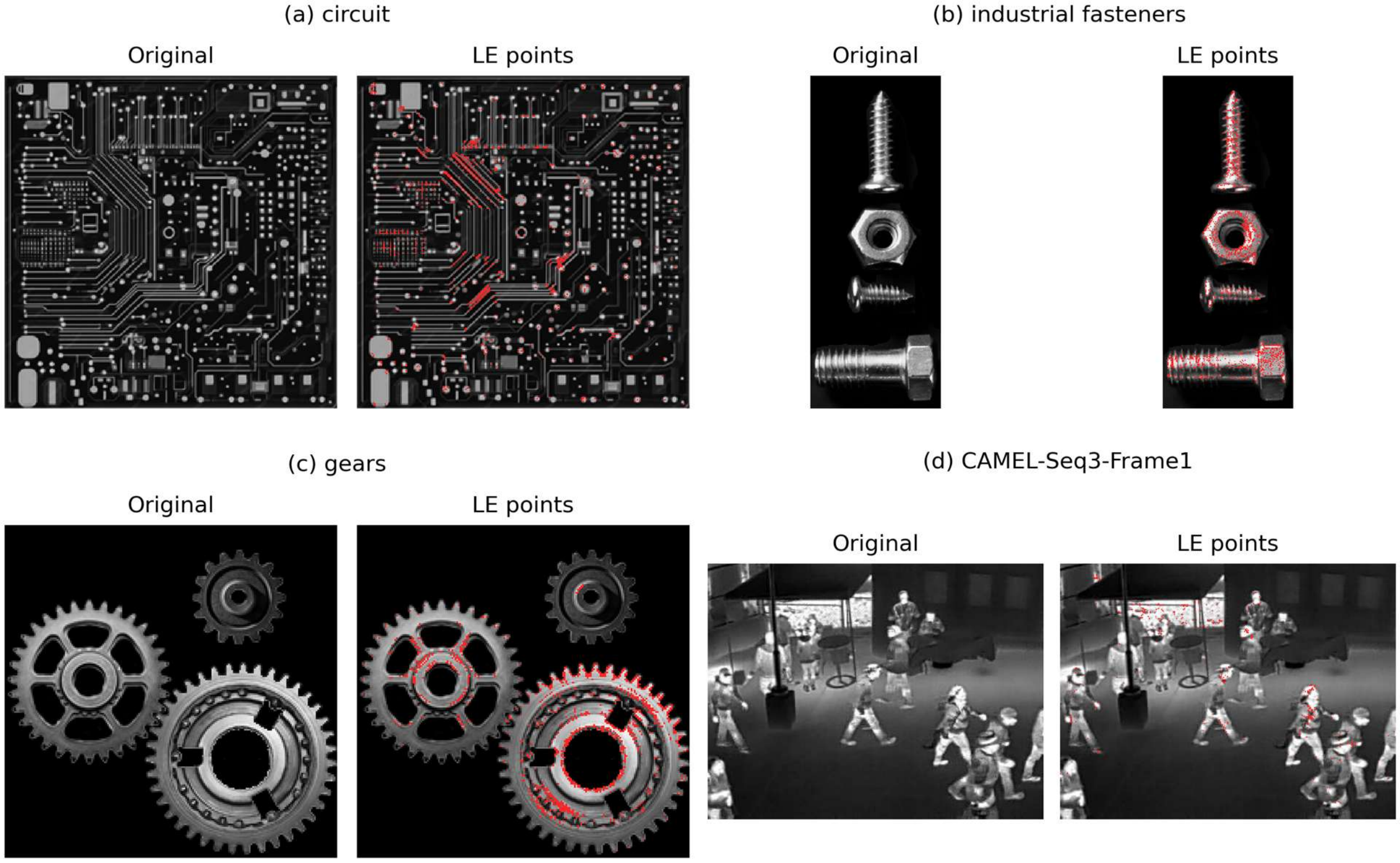

*Fig. 6: Spatial distribution of LE-selected points across the industrial images. The highlighted points denote pixel locations associated with high neural adaptation responses during sample-wise learning, showing structurally informative regions.*

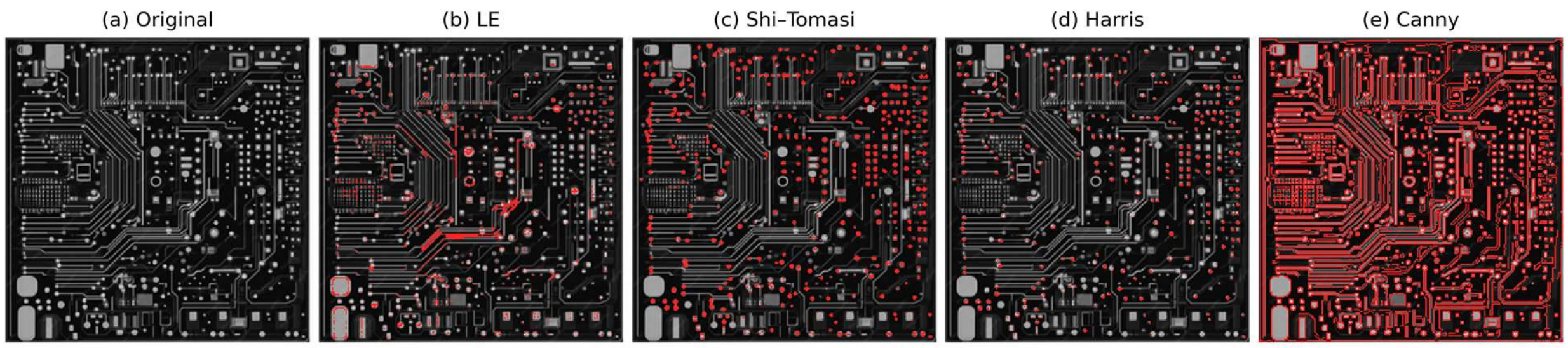

*Fig. 7: Comparison of spatial distributions of points identified by Learning Entropy, Harris, Shi–Tomasi, and Canny methods on the circuit image. LE emphasizes high-adaptation regions, while Harris and Shi–Tomasi identify corner structures, and Canny produces dense edge and contour points.*

## 5. Conclusions

This paper extended the concept of Learning Entropy from temporal adaptive systems toward spatial learning in multilayer perceptrons operating on image data.

By evaluating parameter updates during sample-wise neural adaptation, Spatial Learning Entropy Maps were constructed and used for structural point extraction. The resulting maps reveal image regions that have a strong influence on the adaptation dynamics of the neural model.

Unlike conventional feature extraction approaches operating directly on image gradients, the proposed method evaluates response of a learning system adapting to image structure. The results indicate that adaptation dynamics itself may contain meaningful structural information about image data.

The presented experiments further suggest that adaptation-induced complexity and statistical complexity are fundamentally different quantities. The proposed framework therefore establishes a possible connection between adaptive learning systems, continual learning, and learning-based information measures.